\documentclass[10pt,conference]{IEEEtran}
\usepackage{blindtext, graphicx}
\usepackage{algorithm2e}
\ifCLASSINFOpdf
\else
\fi
\usepackage{multicol}
\usepackage{tabulary,booktabs}

\newcommand\blfootnote[1]{%
  \begingroup
  \renewcommand\thefootnote{}\footnote{#1}%
  \addtocounter{footnote}{-1}%
  \endgroup
}

\begin{document}
%
\title{Deep Learning for System Trace Restoration}



\author{

\IEEEauthorblockN{Ilia Sucholutsky}
\IEEEauthorblockA{
University of Waterloo\\
Waterloo, Canada\\
isucholu@uwaterloo.edu.ca}
\and
\IEEEauthorblockN{Apurva Narayan}
\IEEEauthorblockA{
University of Waterloo\\
Waterloo, Canada\\
apurva.narayan@uwaterloo.ca}
\and
\IEEEauthorblockN{Matthias Schonlau}
\IEEEauthorblockA{University of Waterloo\\
Waterloo, Canada\\
schonlau@uwaterloo.ca}
\and
\IEEEauthorblockN{Sebastian Fischmeister}
\IEEEauthorblockA{University of Waterloo\\
Waterloo, Canada\\
sebastian.fischmeister@uwaterloo.ca}
}



%
\maketitle
\IEEEpeerreviewmaketitle
\begin{abstract}
Most real-world datasets, and particularly those collected from physical systems, are full of noise, packet loss, and other imperfections. However, most specification mining, anomaly detection and other such algorithms assume, or even require, perfect data quality to function properly. Such algorithms may work in lab conditions when given clean, controlled data, but will fail in the field when given imperfect data. We propose a method for accurately reconstructing discrete temporal or sequential system traces affected by data loss, using Long Short-Term Memory Networks (LSTMs). The model works by learning to predict the next event in a sequence of events, and uses its own output as an input to continue predicting future events. As a result, this method can be used for data restoration even with streamed data. Such a method can reconstruct even long sequence of missing events, and can also help validate and improve data quality for noisy data. The output of the model will be a close reconstruction of the true data, and can be fed to algorithms that rely on clean data. We demonstrate our method by reconstructing automotive CAN traces consisting of long sequences of discrete events. We show that given even small parts of a CAN trace, our LSTM model can predict future events with an accuracy of almost 90\%, and can successfully reconstruct large portions of the original trace, greatly outperforming a Markov Model benchmark. We separately feed the original, lossy, and reconstructed traces into a specification mining framework to perform downstream analysis of the effect of our method on state-of-the-art models that use these traces for understanding the behavior of complex systems. 
\blfootnote{\footnotesize \textcopyright 2019 IEEE. Personal use of this material is permitted. Permission from IEEE must be obtained for all other uses, in any current or future media, including reprinting/republishing this material for advertising or promotional purposes, creating new collective works, for resale or redistribution to servers or lists, or reuse of any copyrighted component of this work in other works. \hfill}
\end{abstract}


%

\section{Introduction}
It is often the case that software and algorithms work very differently in lab conditions than they do out in the field. This is particularly a problem for safety-critical systems ranging from planes to pacemakers, where any malfunction in the field can mean serious - even life-threatening -  consequences for users. It is thus important that algorithms designed for such systems work as intended both in the lab and out on the field. The problem is that many such algorithms are developed in the lab using clean, processed, and ‘perfect’ data. Real data, on the other hand, can contain noise and loss that may radically alter the behaviour of the algorithms processing it. This results in these algorithms working in theory, but failing once put into practice due to their dependence on perfect data quality. 

While `imperfect' real data may appear to be irreparable or different from clean data, the reality is that much of the original structure can still be reconstructed from it, especially when there is sufficient information about similar data that can be used for reconstruction. Supervised learning techniques can be used to learn the general structure of similar datasets, and then reconstruct the lossy dataset of interest. 

Many real-world systems produce discrete datasets such as text or event logs during their operation. Automotive Controller Area Network (CAN) traces, for example, consist of a long sequence of discrete events. A CAN bus is a communication system that allows various types of devices like microcontrollers to communicate with each other in real time without needing a host. It is a message-based protocol, designed originally for multiplex electrical wiring within automobiles to save on copper, but is also used in many other contexts. The messages in a CAN trace can be considered as `words' produced by the CAN bus. Therefore, various supervised learning techniques from natural language processing (NLP) are likely to be effective in analyzing the patterns in sequences - or ``sentences"- of these events. 

Recurrent Neural Networks (RNNs), are neural networks that have an internal memory. Long Short-Term Networks (LSTMs) are a type of RNN that have a superior ability to learn long-term dependencies in data. LSTMs have been used extensively for work with NLP and event based data~\cite{seq2seq}, they have been used for data compression~\cite{lstmcompression}, and they have recently even been used - with varying degrees of success - directly for anomaly detection~\cite{lstmanom}. The use of LSTMs for restoration of lossy system traces has not been explored. We propose a method for using LSTMs to accurately reconstruct discrete temporal traces from the CAN bus affected by data loss. The reconstructed traces can then be used by algorithms that rely on `perfect' data. As a case study we will test the performance of a specification mining framework that uses timed regular expressions and deterministic finite state automata for extracting system behavior. 

The model works by learning to predict the next event in a sequence of events from a large database of CAN traces, and uses its own output to continue prediction for multiple points ahead in time, allowing for even large chunks of lost data to potentially be restored. We separately feed the original, lossy, and reconstructed traces into a Timed-Regular Expression Mining (TREM) framework~\cite{Cutulenco:2016,Narayan:2018,Schmidt:2017} to gauge the effectiveness of our LSTM-based reconstruction approach. We show that given even small parts of a CAN trace, the LSTM model can accurately reconstruct large portions of the original trace thereby permitting the use of algorithms like TREM, that are reliant on `perfect' data, with `imperfect' data that would otherwise cause them to perform poorly.

The rest of this paper is divided as follows: Section~\ref{sec:related-work} will discuss related work in data reconstruction. Section~\ref{sec:lstm-architecture} will describe the architecture of the LSTM model and discuss the CAN data used for our experiments. Section~\ref{sec:tre-evaluation} will present a case study with Timed Regular Expression (TRE) mining from system traces, and Section~\ref{sec:results-conclusion} presents results and conclusion.

\section{Related Works}\label{sec:related-work}

\subsection{Deep Learning for Sequence Prediction}
The application of deep learning techniques has led to dramatic success and development of state-of-the-art solutions to numerous real-world problems in computer vision~\cite{deepimagenet}, machine translation~\cite{nmt}, robotics~\cite{oneshot}, etc. In context of sequence prediction, a broad range of techniques have been developed to address the problem found in Natural Language Processing (NLP)~\cite{NLP}, genetic sequencing~\cite{gene}, stock market prediction~\cite{stock}, music generation~\cite{music}, and a whole range of other domains. In particular, RNNs and its subclass, LSTM networks have shown to be very effective for sequence prediction in temporal or sequential data. For example, Gers and Schmidhuber (2001), demonstrated the large improvement in performance by using LSTMs on major language benchmarks used for RNNs~\cite{language}.

Many advances have been made in the closely related problem of using deep learning techniques for generating end-to-end sequences of outputs from a sequence of inputs, beginning with Sutskever et al. (2014), describing a groundbreaking method for using an LSTM-based encoder-decoder system to learn sequences while minimizing the number of assumptions about the structure of the sequence~\cite{seq2seq}. 
Hong et al. (2017), showed that a modified convolutional sequence-to-sequence autoencoder could be trained to predict visually un-observable weather patterns when given preceding satellite images as input~\cite{PSI}; and Marchi et al. (2015) devised a method based on LSTM recurrent denoising autoencoders to predict the features of a consecutive audio frame based on the previous ones~\cite{noise}. 

\subsection{Deep Learning for Lossy Data Recovery}

While numerous studies discuss the use of deep learning for directly solving problems like anomaly detection~\cite{taylor,collective,2017arXiv171009207E,lstmanom} and intrusion detection~\cite{intrusion1,intrusion2, intrusion3}, only very few explicitly focus on the critical intermediate problem of restoring lossy or noisy data that is critical for correct behavior of aforementioned algorithms. Often, missing data are considered unusable and is removed completely in pre-processing to avoid having to address it. When missing data are addressed,  missing values are typically replaced with global or class means~\cite{impintro, climateimp} or neighboring values~\cite{hotdeck, multimp}. These approaches are much too coarse and do not take into account the event sequence or order of occurrence.

Zhou and Huang (2017) discuss this problem and propose a novel LSTM-based approach they call an ``Iterative Imputing Network" for restoring missing sensor data in time-series~\cite{IIN}. While their work is on restoring continuous, multivariate data, we draw inspiration from it and propose a method using LSTMs to restore discrete, sequential, univariate data.

Some other applications of deep learning to data recovery in a variety of domains include: Hsieh and Pratt (2001) show that artificial neural networks could be used to recover lossy field data with high reliability~\cite{hsieh_pratt}; Tilk and Alum\"{a}e (2015) show that LSTMs are effective at restoring punctuation in unpunctuated streams of text~\cite{speech}; and Haque, Yousuf, and Rana (2018) show that a combination of CNNs and LSTMs could be used to successfully de-noise and restore image data~\cite{CNN-LSTM}.

\section{LSTM}\label{sec:lstm-architecture}

\subsection{RNNs and LSTMs}
One of the major limitations of non-recurrent neural networks is that they operate on fixed-size vectors, performing a limited number of transformations to derive another fixed-size vector. This limits the effectiveness of non-recurrent networks in identifying features dispersed over sequences or over time. Recurrent neural networks (RNNs) are a type of neural network that use a feedback mechanism to allow the network to operate on a sequence of inputs as well as outputs by selectively keeping information about previous states. In this manner, the $i^{th}$ output vector \(y_i\) of an RNN layer is a function of the $i^{th}$ input \(x_i\) to the layer as well as the previous output \(y_{i-1}\) of the layer. As with regular layers, the function is generally a non-linear transformation over a weighted sum of the terms involved.

\[y_i = f(y_i, x_i, y_{i-1}) = f(V*y_{i-1} + W*x_i + c)\]

Of course in such a recursive function, every previous output and input must be considered when adjusting weights during training, as during backpropagation they would all have an impact on the error gradient. These incredibly long chains of derivatives pose two problems: computational intensity and vanishing or exploding gradients. 

\begin{figure}[h!]
\caption{\textbf{Single RNN Unit:} A recurrent neural network uses a feedback mechanism to access information about previous states.}
\vspace*{5mm}
\centering
\includegraphics[scale=0.4]{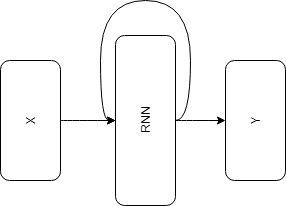}
\label{Single}
\end{figure}

\begin{figure}[h!]
\caption{\textbf{Unrolled RNN Unit:} The feedback loop in a recurrent neural network can be unfolded for an alternative, sequential representation of the repeated transformations it performs.}
\centering
\includegraphics[scale=0.4]{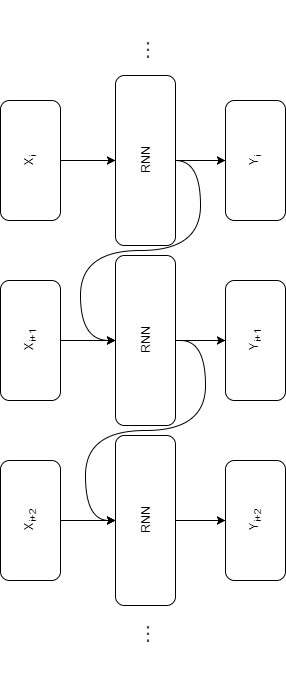}
\vspace{-10 pt}
\label{Unroll}
\end{figure}

One solution to these problems is known as \textbf{unrolling} the RNN. This requires considering that if an RNN is essentially a loop with signal flowing through, it can also be equally represented by `unfolding' the loop as a long series of the same transformations applied to the signal. The structures in Figure~\ref{Single} and Figure~\ref{Unroll} are thus equivalent ways of representing an RNN. Unrolling is the process of using the second, `unfolded' representation of an RNN and having the option to consider only a fixed number of previous states when predicting the current output. If we use the analogy of short-term memory to describe the feedback mechanism of an RNN, then unrolling would be displaying this memory as a sequence of events and having the option to cut off the sequence at some event, ignoring any events that came before it.

A second solution to the problem of vanishing or exploding gradients in recurrent networks was the creation of two gates: input and forget. The input gate controls which information from the current state's input will be used, while the forget gate controls which information will be used from the recurrent outputs of previous states. In such a way, the input and forget gates allow the network to \textbf{selectively} remember or forget information about previous and current states. Hochreiter and Schmidhuber (1997) proposed a new type of RNN that makes use of these two gates in their recurrent units to solve the vanishing gradient problem and gave it the name Long Short-Term Memory network (LSTM)~\cite{LSTM}.

\subsection{Architecture}
\begin{figure}[h]
\caption{\textbf{LSTM Model Architecture:} The LSTM model consists of two hidden layers followed by two recurrent LSTM layers and one additional hidden layer. Input is a sequence of events, output is a prediction of next event in the sequence. Loss is calculated as a logloss function comparing the true next event to the predicted one.}
\vspace*{8mm}
\centering
\includegraphics[scale=0.4, angle=270]{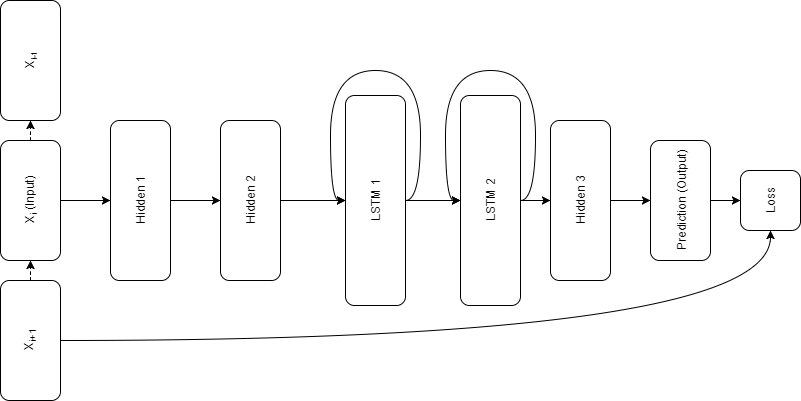}
\label{Basic}
\end{figure}

We built our LSTM model in Tensorflow. After experimenting with a few different architectures, we found that a smaller model with five layers  in the hidden portion as seen in Figure \ref{Basic} was sufficient.  Additional hidden layers did not improve performance.

The input layer has one node for each entry in the trace dictionary, and one additional ``other" node to account for rare events that were not seen in the training data that was used to compile the dictionary. Input dropout was set to 0.2. 

It is followed by two densely-connected hidden layers with double the number of nodes as in the input layer. Each of these hidden layers uses a hyperbolic tangent activation function, and has a dropout of 0.4. 

The hidden layers are followed by 2 LSTM layers, using Tensorflow’s LayerNormBasicLSTMCell. These cells perform layer normalization based on Ba, Kiros, and Hinton (2016)~\cite{layernorm}. Each one is unrolled for 40 steps; increasing unrolling further did not seem to increase performance on this dataset, but may be useful for others such as those with higher dimensionality. Both LSTM layers also use the hyperbolic tan activation function, and recurrent dropout of 0.4 based on Semeniuta, Severyn, and Barth (2016)~\cite{recdrop}. Each has four times as many nodes as each of the hidden layers, or eight times as many as the input layer.

Finally, the last hidden layer is another densely connected layer with a sigmoid activation function and our final configuration has the same number of nodes as the input layer. However, in an earlier configuration explained below, this layer had n times the number of nodes as the input layer, where n is the number of future events the user wishes to simultaneously predict. The output of this layer is the output of the model.

To increase training speed, LayerNormBasicLSTMCell layers could be replaced with either the peephole~\cite{peephole} or the non-peephole~\cite{LSTM} implementation of CoupledInputForgetGateLSTM layers that couple the input and forget gate as described by Greff et al. (2015)~\cite{coupled} resulting in less computational operations but higher variance in performance. For our CAN trace experiment, the vocabulary size was 43, our input layer had 44 nodes, our hidden layers had 88 nodes each, our LSTM layers had 352 nodes each, and the last hidden layer had 44 nodes whose output was considered the output of the model.

\subsection{Data}\label{data}
Before combining the LSTM approach with the Timed Regular Expression Mining (TREM), it was necessary to test whether the LSTM approach works at all. To do so, a simple dataset was needed that was guaranteed to have temporal patterns but is not trivial. Many messages in automotive CAN data occur in a periodic way so temporal patterns are present; however, other messages in CAN data are event-triggered and thus predicting upcoming messages is a non-trivial problem. To keep the dataset simple, we do not attempt to predict each bit of the CAN message payload but only each message ID.  

Our CAN data comes from a Lexus RX450h hybrid SUV. The data are split into a number of maneuvers that were repeated multiple times. To reduce some of the variance and reduce training time while testing our LSTM approach, we used traces from a single maneuver, the vehicle driving at 20 km/h and decelerating down to 0 km/h. 20 clean traces of this maneuver were available. 

Of the 20 traces, 15 were used for training and validation, while 5 were held out for testing. The traces used for training were examined to compose a dictionary of possible message IDs. A total of 43 different message IDs were found in the training set. We added a 44th element to the dictionary to designate any “other” message IDs that may not have been present in the training set. We then one-hot encoded all  traces using this dictionary, replacing the one-dimensional message ID with a 44-dimensional vector of indicator variables where 43 elements have the value 0 and 1 element the value 1.   

\subsection{Training}
The LSTM was trained using the following procedure:
\begin{enumerate}
    \item Select 2 random traces from set (1 for training, 1 for validation)
    \item Train the model on these 2 traces for 10 epochs with a learning rate of 0.2 and a logloss loss function
    \item Train the model on these 2 traces for an additional 20 epochs with learning rate decay of 1/1.1 and a logloss loss function
    \item Reset learning rate, preserve weights, repeat from step 1 selecting 2 new random traces
\end{enumerate}
Dropout for regularization and randomized input order prevent overfitting, so this procedure should be repeated until accuracy reaches a plateau since early termination due to detection of overfitting is unlikely to occur. Alternatively, the procedure can be repeated a fixed number of times if so desired.

\subsection{Benchmark}
In order to understand what the results mean, it is important to have a benchmark. We create a benchmark using a fast Markov Model that could be described as a history search, or as a conditional probability method. It has been shown that in situations where data or computational power are limited, Hidden Markov Models can match the performance of LSTMs~\cite{lstmvshmm}. As a result, the Markov Model will act as a benchmark that the LSTM can be compared against. 

The basic premise of this method is that the next message is highly correlated with what the preceding messages were in a sequence. However, this correlation is likely to degrade as we look further into the past. As a result, we can say that the system is a Markov process of order $n$, where the previous $n$ messages form a state that influences the next one. Sequences of $n$ consecutive messages are often called ``n-grams'', and their analysis is common in sequence modelling domains like Natural Language Processing (NLP)~\cite{ngram, ngram2, ngram3}.  
The most straightforward method of using this property is to perform a history search where every time we want to make a prediction, we look at the previous $n$ messages, and then search our entire training dataset to find the most commonly occurring message after this n-gram. This of course, can involve performing a large number of repeated searches since we do not store the results between searches.

Instead we can  build a model of the system by selecting an $n$, and creating a separate state for each possible combination of the $d$ unique messages. Of course, initializing all possible states at once means that our model will contain $n^d$ states even if some of these are never seen in the data, which may quickly become too memory-intensive for large $n$ or $d$. To mitigate this issue, we iteratively fill a dictionary $D$ by traversing the training data using the following algorithm:
\begin{algorithm}
\caption{Iteratively learn transition frequencies}
\SetAlgoLined
\KwResult{Dictionary $D$ containing all states and transition frequencies found in the training data }
$i=0$\;
\While {$i + n < length(train\_data)$}{
    $K=train\_data[i:i+n]$\;
    \If {$K \notin D.keys()$}{  
    Let $S_K$ be a new dictionary\;
    \For{each $m_i$ in the set of the $d$ unique messages}{
        $S_K[m_i]:=0$\;}
    $D[K] := S_K$\;}
    $m^* = train\_data[i+n]$\;
    $S_K[m^*]=S_K[m^*]+1$\;
    $i=i+1$\;}
\end{algorithm}

During inference, each time we would like to impute a value, we take the $n$ preceding values to create a state and look it up in the dictionary to find the highest probability transition. If there are multiple missing values within $n$ steps of each other, then we impute them chronologically and use our previously imputed values when imputing the consecutive ones. One issue that arises for larger $n$ and $d$ or smaller training datasets, is that there may be states in the testing data that are not found in the training data. To address this, we updated $D$ to include sub-sequences of length less than $n$ as states; $D$ now contains k-grams where $k\leq n$. When we want to impute a missing value, we find the maximal length state in $D$ that matches the preceding values. 

After experimenting with different values for the hyper-parameter $n$, we found that accuracy reached its peak and stayed constant above $n\geq 30$; for consistency with the LSTM experiments we used $n=40$. The maximal next-message prediction accuracy achieved with this Markov model was 76.55\%, but when predicting multiple messages consecutively, the accuracy rapidly dropped off, falling to just 37.01\% when predicting 20 events into the future. Tables in the Results section detailing accuracy of the LSTM method also include benchmark results. 
\subsection{Results}
Accurately predicting multiple steps forward is a challenging but important problem in lossy data restoration, as multiple consecutive events may have been lost. Two ways of predicting multiple events forward were considered. 

The first method was to increase the size of the output layer by a factor of n to directly predict n events forward. In other words, the output layer was modified to have n*44 nodes where n was number of events to simultaneously predict. The output at each step was reshaped into n vectors of 44 nodes each, and each vector predicted 1 future event. Within each vector, the element with the highest value was considered as the prediction. The true positive rate was used as a measure of accuracy. A set of predictions at step i was considered to be a true positive if all n of n predictions made at that step were made correctly. If \(c_i\) denotes the number of steps in epoch i where all n predictions were correct, and \(w_i\) denotes the number of steps in epoch i where at least 1 of the n predictions was incorrect, then the accuracy for that epoch is denoted by:
\[acc_i = \frac{c_i}{c_i+w_i}\]

One of the issues with this setup is that each time n is changed, a new network needs to be initialized and trained. As such, we retrained the model several times with increasingly large values of n. Table \ref{direct} documents the average performance of the model for different values of n as well as the range 95\% of the measured accuracy levels were within. The average accuracy with this method decayed rapidly as n was increased, while the ranges increased quickly. This suggests that the random initialization of the network plays an increasingly large role in the quality of predictions as n increases. 

\begin{table}[h]
\caption{n-forward prediction accuracy using the direct method}
\label{direct}
\begin{center}
\begin{tabular}{ |c|c|c|c| } 
 \hline
 n & Avg. accuracy & 95\% range & Benchmark \\
 \hline
 1 & 0.895 & 0.88-0.91 & 0.766\\ 
 10 & 0.58 & 0.52-0.65 & 0.454\\ 
 20 & 0.48 & 0.4-0.54 & 0.370\\
 \hline
\end{tabular}
\end{center}
\end{table}

A second method was developed to address the shortfalls of the first. The second method predicts only one output at a time; however, the code was altered to allow the model to use its predictions as inputs to itself. In such a way, the model uses its own outputs, one at a time, to make long sequences of predictions. Using the numbers from Table \ref{direct}, the expected accuracy for this method when predicting n steps forward would be the accuracy for 1 step prediction, to the power of n, under the assumption that the model's predictions derail after it makes even a single mistake. Table \ref{indirect} details the expected accuracy levels for this second method based on this calculation. 

\begin{table}[h]
\caption{Expected n-forward prediction accuracy using the step-by-step method assuming model cannot recover after a mistake}
\label{indirect}
\begin{center}
\begin{tabular}{ |c|c|c| } 
 \hline
 n & Expected accuracy & Benchmark\\
 \hline
 1 & 0.895 & 0.766\\ 
 10 & 0.330 & 0.454\\ 
 20 & 0.109 & 0.370\\
 \hline
\end{tabular}
\end{center}
\end{table}

\begin{table}[h]
\caption{True n-forward prediction accuracy using the step-by-step method}
\label{true}
\begin{center}
\begin{tabular}{ |c|c|c| } 
 \hline
 n & Accuracy & Benchmark\\
 \hline 
 1 & 0.895 & 0.766 \\  
 10 & 0.892 & 0.454\\ 
 20 & 0.889 & 0.370\\
 \hline
\end{tabular}
\end{center}
\end{table}
While the expected accuracy values for this method were very low, the convergence in logloss in the model suggested that this accuracy estimate may be underestimating model performance. In fact, when the model was run on the test traces using this step-by-step method, the true results were drastically different as shown in Table \ref{true}. This suggests that even if the model makes a mistake in its predictions, it is robust enough to continue predicting correctly contrary to the assumption above. However, it is clear that the assumption does hold for the benchmark model and as a result it is not nearly as robust to its own mistakes as the LSTM is. 

As mentioned above, the LSTM uses 40 steps of unrolling for the recurrent portion meaning the model requires 40 inputs to fill its internal feedback sequence. Impressively, when given just the first 40 events of the shortest CAN trace held out for testing, the model was able to step-by-step predict the remaining 3500 events of this trace with only several omissions of rarely occurring events as in Table \ref{omitted} and some localized mistakes in the order of predicted events as in Table \ref{ordering}. 

\begin{table}[h]
\caption{\textbf{Example of proper alignment:}} {All events in this segment were predicted correctly and match their true counterparts.}
\label{propalign}
\begin{center}
\begin{tabular}{ |c|c c| } 
 \hline
 & Predicted Event & True Event\\
 \hline
1 & 2C6 & 2C6\\
2 & 5D7 & 5D7\\
3 & B0 & B0\\
4 & 224 & 224\\
5 & B2 & B2\\
6 & 20 & 20\\
7 & B4 & B4\\
8 & 25 & 25\\
9 & 22 & 22\\
10 & 23 & 23\\
 \hline
\end{tabular}
\end{center}
\end{table}

\begin{table}[h!]
\caption{\textbf{Example of omitted rare event:}}{Rarely occurring event `340' was incorrectly omitted by the model, causing all predicted events beginning from the fourth one to be shifted one up from their true counterparts.}
\label{omitted}
\begin{center}
\begin{tabular}{ |c|c c| } 
 \hline
 & Predicted Event & True Event\\
 \hline
1 & B4 & B4\\
2 & 25 & 25\\
3 & 22 & 22\\
4 & 23 & 23\\
5 & B0 & \textbf{340}\\
6 & 320 & B0\\
7 & B2 & 320\\
8 & 2D0 & B2\\
9 & 2C4 & 2D0\\
10 &  & 2C4\\
 \hline
\end{tabular}
\end{center}
\end{table}

\begin{table}[h!]
\caption{\textbf{Example of local ordering mistake: }} {'2c4' was incorrectly predicted as the 8th event instead of 4th, causing all of the other events from 4th to 8th to also appear misclassified. In reality, true events 5-8 were shifted up by one and predicted as events 4-7.}
\label{ordering}
\begin{center}
\begin{tabular}{ |c|c c| } 
 \hline
 & Predicted Event & True Event\\
 \hline
1 & 25 & 25\\
2 & 22 & 22\\
3 & 23 & 23\\
4 & 2C6 & \textbf{2C4}\\
5 & B0 & 2C6\\
6 & 320 & B0\\
7 & B2 & 320\\
8 & \textbf{2C4} & B2\\
9 & 20 & 20\\
10 & 223 & 223\\
 \hline
\end{tabular}
\end{center}
\end{table}

\begin{figure}[t]
\caption{\textbf{One-hot Encoded True Events:} Visualization of a sequence of just over 100 true events pulled from a testing trace. White pixels correspond to the one active element in that column.}
\vspace*{8mm}
\centering
\includegraphics[scale=0.6]{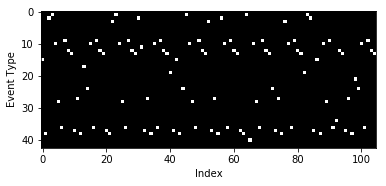}
\label{t_embed}
\end{figure}

\begin{figure}[t]
\caption{\textbf{One-hot Encoded Predicted Events:} Visualization of a sequence of just over 100 predicted events pulled from the predictions on a testing trace. White pixels correspond to the one active element in that column.}
\vspace*{8mm}
\centering
\includegraphics[scale=0.6]{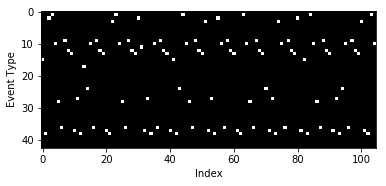}
\label{p_embed}
\end{figure}

Figures \ref{t_embed} \& \ref{p_embed} are visualizations of a sequence of true one-hot embedded events and a sequence of predicted one-hot embedded events, respectively, pulled from the same portion of a test trace. It is noticeable that mismatches between the two Figures get increasingly worse as the index increases. This is due to the omission problem: each time an event is omitted in the predictions, the entire sequence of predicted events is shifted to the left, causing increasingly large mismatches. When gauging model performance, each omission was recorded, and then the predicted sequence re-aligned at that point in order to once again match the true sequence. Similarly, each ordering mistake was recorded, and the one point detected to be in the wrong position moved to its correct location, to identify whether other mistakes were made in the same area. 

On average, the model omitted 5.058\% of points and had one local ordering mistake every 11.1 events. Curiously, both these mismatches and ordering mistakes occur at only a slowly increasing rate throughout all of the 3500 predictions, suggesting once again that our model's flexibility makes it at least partially resistant to mistakes in the input, as prediction quality did not decrease even if some local mistakes in output prediction were made and then fed in as input. The LSTM-based approach is thus sufficiently robust to work with and restore not only lossy, but also noisy, data.

\section{Case Study: Timed Regular Expression Mining}\label{sec:tre-evaluation}

Time of occurrence of events or tasks in real-time systems is critical to the correct operation of these systems~\cite{lamport1978time,dwyer1999patterns}. Software for these systems have become quite complex and it is challenging for engineers to understand the underlying behavior of the systems for various tasks such as debugging, root-cause analysis, etc. Mining software specifications from the system traces for these systems plays a key role in understanding the underlying interactions in these systems~\cite{lemieux2015general}. It becomes more important due to unavailability of clear software specifications. These are temporal specifications in the context of real-time systems and are used in various tasks such as software testing~\cite{dallmeier2010generating}, verification~\cite{kincaid2015automated}, etc. 

In real-time systems, the temporal specifications are required to address not only the qualitative notion of time (i.e. ordering of events), but also the quantitative notion of time (i.e. time of occurrence of event). Processes in safety-critical systems are required to adhere to strict timing constraints and deadlines, violating them can lead to catastrophic outcomes. Most temporal specification mining frameworks are based on the idea of state machines. These temporal specifications can be used for numerous tasks such as anomaly detection, run-time monitoring, etc. 

We present a use-case based on the temporal specification mining framework by Narayan et. al. (2018)~\cite{Narayan:2018}. The framework takes a TRE template and a system trace as input and computes the dominant set of properties (a set of most commonly occurring rules) in the form of TRE instances. The framework builds upon the method proposed by Asarin et. al. (2002)~\cite{timedregex} for synthesizing of a timed automaton. 

The following two TRE templates are used for the evaluation of traces from a real car as explained in Section~\ref{data}. The time interval was standardized between 0 to 1,000. We executed the TRE mining algorithm on all three types of traces for the purpose of evaluation: Normal, Lossy, and Recovered. We used two rules (alternating and response pattern~\cite{dwyer1999patterns}) that represent common behavior of real-time systems. The templates of the rules are in the form of TRE as shown below, where P and S represent unique events in system traces, and *, $|$, $+$ and $.$ are operators:\\

{
\noindent \textbf{T-1(response)}: $(\hat{}( P)^{*}.(\langle P. \, \hat{}(S)^{*}.S\rangle[0,1000]).\,\hat{}(P)^{*})+$ 

\noindent \textbf{T-2(alternating)}: $(\hat{}(P|S)^*.(\langle P.\,\hat{}(P|S)^*.S.\hat{}\,(P|S)^*\rangle [0,1000]))+\\$
\par}

The mining framework extracts all rules from the system traces that take the form of the above two templates. A ranking module reduces the mined set of rules to a set of most commonly occurring TRE-instances in the system trace. We compared the presence and absence of the mined TRE instances from normal, lossy, and restored traces for evaluation of our restoring algorithm.

We run TREM on original traces to get the true number of mined instances. Then we run TREM on lossy and restored traces that have a controlled fraction of messages lost. The results of specification mining are presented in Table \ref{tab:tre-mine-results} where we show the percent of original TRE instances that are \textbf{not} found when mining lossy and restored traces at varying levels of loss in the trace. A higher quality restoration results in more of the original TRE instances being mined from the restored trace (ie. a perfect restoration would result in a decrease of 0\% from original to restored).

\begin{table}[h]
\caption{Percent decrease in number of mined TRE instances at each level of loss compared to the number mined in original traces}
\label{tab:tre-mine-results}
\begin{center}
\begin{tabulary}{\linewidth}{ |c|c c c c c| } 
 \hline
\textbf{Message Loss} & \textbf{5\%} &  \textbf{10\%} & \textbf{15\%} & \textbf{20\%} & \textbf{25\%} \\
 \hline
\textbf{Lossy Traces} & 6.2\%  & 17.8\% & 21.6\% & 33.37\%  & 53.99\%   \\
 \textbf{Restored Traces} & 5.9\% & 8.2\% & 9.8\% & 12.6\% & 10.8\% \\ 
 \hline
 \textbf{Improvement} & 0.3\% & 9.6\% & 11.8\% & 20.77\% & 43.91\% \\
 \hline
\end{tabulary}
\end{center}
\end{table}

Clearly, performing the LSTM restoration improves the mining performance greatly.
When the traces are lossy, the number of TRE instances found are reduced. When the traces  are restored using our LSTM model, the number of TRE instances found is still reduced but much less so, particularly when the percentage of loss is large.
For example, when 25\% of the messages are lost,  performing the LSTM restoration leads to a loss of only 10.8\% of the TRE-instances as compared to 53.99\% of the instances without restoration.

\section{Conclusion}\label{sec:results-conclusion}
Clean data are a crucial component of training and running numerous machine learning algorithms. Unfortunately, data in the field are often contaminated with noise, loss, and other imperfections. We have developed an LSTM-based approach for restoring lost or noisy data in discrete settings such as CAN traces. This approach can be used to restore or predict arbitrarily large sequences of missing data, using its output as input to predict further into the future. A major advantage of this approach is that few assumptions need to be made about the structure of the data, and so it can be applied in any setting where there is discrete data with some form of temporal or sequential dependence. 

While the performance of this method was high in the settings we tested it in, there are certainly changes that can be made to further improve it. For example, more work needs to be done to ensure accuracy is just as high when the model is not trained on just a single maneuver but rather on normal driving traces where maneuvers may not be known at time of prediction. 

A potential method for increasing the accuracy of lossy data restoration would be using context not only from events preceding the lossy portion, but also from events after it. A bidirectional RNN could be trained to restore lossy sections of data when given some number of events from both before and after those sections. The aforementioned IIN method developed by Zhou and Huang (2017) does just that, using a forward and backward LSTM to learn the structure of a time series~\cite{IIN}. However, the advantage of using a one-directional LSTM as in our implementation is that once the model is trained, lossy data can be recovered effectively in real-time while a bidirectional recurrent method would require some latency to allow new events to come in.

A second method for increasing accuracy and decreasing dependence on separation by maneuver, would be improved regularization within the network. In particular, zoneout, which stochastically preserves hidden activations instead of dropping them entirely, has recently been shown to be a more effective method for recurrent regularization than recurrent dropout~\cite{zoneout}. 

Our solution of using one-hot encoding and having an ``other" node in the dictionary of events works well for data where a majority of event types are known in advance and are consistent between training and test data. Another direction worthy of exploration would be finding a method to avoid having to pre-compile a dictionary of events. This is especially crucial in systems where new events are synthesized, or there are a large number of rare events. One solution could be to use a different embedding for the input data, that can be applied to new events on the fly. 

A method that may be able to solve both the issues of performing well on multiple maneuvers and of embedding the input, could be to make use of an encoder network that would take data as input, perform transformations to embed the data in a different space, and send its output to the LSTM network. This would be more consistent with the sequence-to-sequence networks described by Sutskever et al. (2014)~\cite{seq2seq} and we must compare our method with these approaches to determine whether the change in performance merits the additional complexity. 

Finally, for training, it is likely that using AdamOptimizer as described in Kingma and Ba (2014) will lead to faster convergence~\cite{adam} than using GradientDescentOptimizer with decaying learning rate as we did with our LSTM. A few limited runs with a smaller model appear to confirm this, but more testing is needed to determine whether the effect is significant.

Our LSTM-based method is nonetheless an effective standalone solution for recovering lost discrete data in the field. It can be used to pre-process data intended to be used with any other algorithm in order to improve end-to-end performance. 

\clearpage


%







%

\bibliographystyle{unsrt}
\bibliography{biblio}

%





\end{document}